\documentclass[10pt,twocolumn,letterpaper]{article}

\usepackage[pagenumbers]{cvpr} 

\def\name{PVC}
\definecolor{mygray}{gray}{0.7}
\newcommand{\demph}[1]{\textcolor{mygray}{#1}}

\newcommand\blfootnote[1]{
    \begingroup
    \renewcommand\thefootnote{}\footnote{#1}
    \addtocounter{footnote}{-1}
    \endgroup
}

\definecolor{cvprblue}{rgb}{0.21,0.49,0.74}
\usepackage[pagebackref,breaklinks,colorlinks,allcolors=cvprblue]{hyperref}
\usepackage{multirow}
\usepackage{makecell}
\usepackage{mathtools}
\usepackage[numbers]{natbib}
\usepackage{marvosym}

\newcommand*{\affaddr}[1]{#1} 
\newcommand*{\affmark}[1][*]{\textsuperscript{#1}}

\def\confName{CVPR}
\def\confYear{2025}

\title{PVC: Progressive Visual Token Compression for \\Unified Image and Video Processing in Large Vision-Language Models}

\author{
\textbf{
Chenyu Yang\affmark[1,3$*$],\:
Xuan Dong\affmark[1,3$*$],\:
Xizhou Zhu\affmark[1,2$*$],\:
Weijie Su\affmark[2$*$],\:
Jiahao Wang\affmark[3],\:
Hao Tian\affmark[3],}\\
\textbf{Zhe Chen\affmark[2,4],\:
Wenhai Wang\affmark[2,5],\:
Lewei Lu\affmark[3],\:
Jifeng Dai\affmark[1,2\Letter]} \\[1mm]
\affaddr{\affmark[1]Tsinghua University}\quad
\affaddr{\affmark[2]OpenGVLab, Shanghai AI Laboratory}\quad
\affaddr{\affmark[3]SenseTime Research}\\
\affaddr{\affmark[4]Nanjing University}\quad
\affaddr{\affmark[5]The Chinese University of Hong Kong} \\
\texttt{\small \{yangcy23,x-dong21\}@mails.tsinghua.edu.cn,}
\texttt{\small \{zhuxizhou,daijifeng\}@tsinghua.edu.cn,} \\
\texttt{\small suweijie@pjlab.org.cn,}
\texttt{\small \{wangjiahao2,tianhao2,luotto\}@sensetime.com,} \\
\texttt{\small chenzhe98@smail.nju.edu.cn,}
\texttt{\small whwang@ie.cuhk.edu.hk} \\[1mm]
\small Code: \url{https://github.com/OpenGVLab/PVC}
\vspace{-2em}
}

\begin{document}
\maketitle

\blfootnote{$^{*}$Equal contribution. This work is done when Chenyu Yang and Xuan Dong are interns at SenseTime Research.\textsuperscript{\Letter}Corresponding author: Jifeng Dai \textless daijifeng@tsinghua.edu.cn\textgreater.}

\begin{abstract}
Large Vision-Language Models (VLMs) have been extended to understand both images and videos. 
Visual token compression is leveraged to reduce the considerable token length of visual inputs.
To meet the needs of different tasks, existing high-performance models usually process images and videos separately with different token compression strategies, limiting the capabilities of combining images and videos.
To this end, we extend each image into a ``static'' video and introduce a unified token compression strategy called Progressive Visual Token Compression (PVC), where the tokens of each frame are progressively encoded and adaptively compressed to supplement the information not extracted from previous frames.
Video tokens are efficiently compressed with exploiting the inherent temporal redundancy. Images are repeated as static videos, and the spatial details can be gradually supplemented in multiple frames.
PVC unifies the token compressing of images and videos. With a limited number of tokens per frame (64 tokens by default), spatial details and temporal changes can still be preserved.
Experiments show that our model achieves state-of-the-art performance across various video understanding benchmarks, including long video tasks and fine-grained short video tasks. 
Meanwhile, our unified token compression strategy incurs no performance loss on image benchmarks, particularly in detail-sensitive tasks.
\end{abstract}
\vspace{-1.5em}

\section{Introduction}
\label{sec:intro}

\begin{figure}[t]
    \centering
    \includegraphics[width=1.0\linewidth]{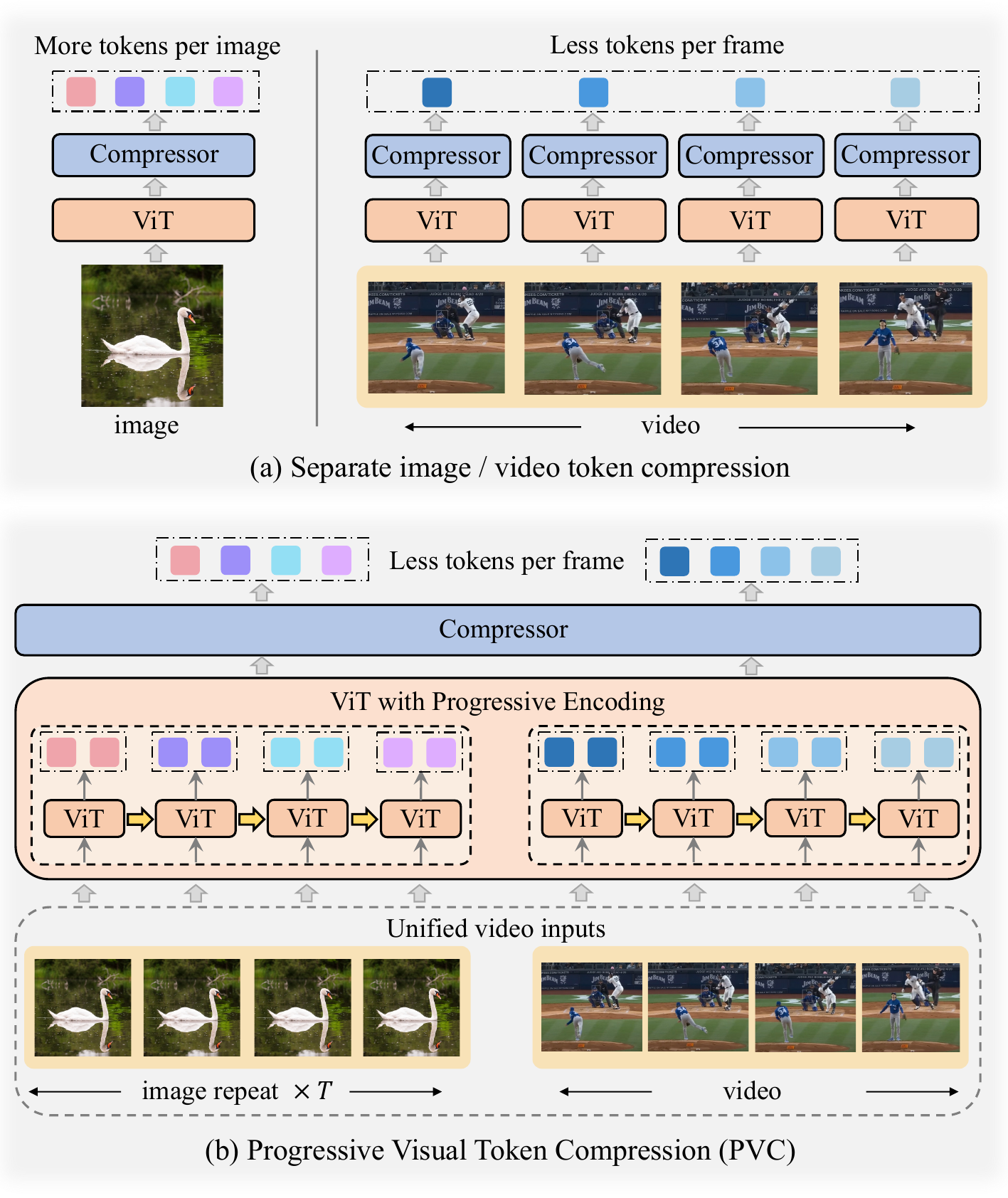}
    \caption{\textbf{Comparison of token encoding and compression in VLMs.} \textbf{(a)} Existing VLMs compress image and video tokens separately, leading to inconsistency: more tokens per image benefit image spatial perception, while videos tend to sacrifice some tokens per frame to accommodate more frames. \textbf{(b)} Our progressive compression (\name{}) achieves unified compression of images and videos, allowing for the continuous supplementation of image details and temporal dynamic information in subsequent frames.}
    \label{fig:teaser}
    \vspace{-1.5em}
\end{figure}

Recently, large Vision-Language Models (VLM)~\cite{liu2024visual,liu2024improved,li2024llava,chen2024internvl,chen2024far,internvl2,bai2023qwen,wang2024qwen2,gpt4o,gpt4v,sonnet3_5,gemini} have achieved remarkable success in various image-text tasks, demonstrating strong general capabilities. Many studies have further extended these models to handle both images and videos. While videos can be seen as a natural extension of images over time, the additional temporal dimension poses new challenges. VLMs typically use visual tokens to represent visual information. A single image often requires hundreds or even thousands of tokens. If each frame of a video is encoded as an image in the same way, a 100-frame video would generate over tens of thousands of tokens, causing a significant computational burden during training and inference.
To address this, VLMs use a technique called \textit{visual token compression}~\cite{wang2024qwen2,chen2024far,li2024llava} to reduce the number of tokens and lower computational complexity. The key challenge is finding a way to reduce the number of tokens while retaining essential information.

As shown in Fig.~\ref{fig:teaser} (a), many high-performance VLMs usually process images and videos separately, using different visual token compression methods to meet the specific needs of each. For images, the models need to preserve fine-grained spatial details to support detail-intensive image tasks (e.g., OCR). For videos, current tasks often focus more on capturing dynamic changes over time and require fewer spatial details. This allows for fewer tokens per frame, sacrificing some spatial details to accommodate more frames. For example, Qwen2-VL~\cite{wang2024qwen2} and LLaVA-OneVision~\cite{li2024llava} reduce the number of tokens per frame for videos compared to tokens per image to handle more frames. However, treating images and videos separately can hinder their combined capabilities, such as limiting the ability of video to recognize spatial details.

An ideal visual token compression method should support unified image and video processing, effectively capturing spatial details and temporal changes. To achieve this, as illustrated in Fig. \ref{fig:teaser} (b), we extend each image into a static video and introduce the concept of ``progressive compression'' for videos: the current frame only encodes new information not extracted from previous frames. This effectively reduces representation redundancy and lowers the number of tokens required for each frame.
For the \textit{video} modality, adjacent frames are often similar and naturally have temporal redundancy. Encoding each frame independently results in repeated information and wasted tokens. By applying progressive compression, token redundancy is minimized, and temporal changes can be captured efficiently.
For the \textit{image} modality, a single image is repeated into multiple frames to form a static video. Progressive compression can then be applied to these frames, allowing spatial details to be gradually supplemented in subsequent frames.
This approach unifies the processing of images and videos. Even with a limited number of tokens per frame,  spatial details and temporal changes can be preserved.

Specifically, our proposed \emph{\textbf{P}rogressive \textbf{V}isual Token \textbf{C}ompression (PVC)} strategy is achieved with the following architectural designs:
(1) We standardize VLM inputs as videos. Each image is repeated multiple times (4 times by default) to form a ``static'' video. 
(2) A \emph{progressive encoding module} is introduced to the vision encoder. 
This module consists of a causal temporal attention to avoid redundant information and progressively extract information complementary to the representations of previous frames. An AdaLN operator injected with timestep information is adopted to differentiate different frames and avoid redundant encoding.
(3) An \emph{adaptive compression module} is introduced to compress the visual tokens without redundancy. 
Based on PixelShuffle~\cite{shi2016real} with shared MLP (i.e., the token compression method used by the open-source VLM InternVL~\cite{chen2024far,internvl2}), we incorporate an additional AdaLN operator to extract different spatiotemporal information at different time steps. 

We evaluate our model on a wide range of multi-modal benchmarks. 
Our model achieves state-of-the-art results across various video understanding benchmarks, including long video tasks (e.g., VideoMME~\cite{fu2024video}, MLVU~\cite{zhou2024mlvu}) and fine-grained short video tasks (e.g., MVBench~\cite{li2024mvbench}).
Meanwhile, on image understanding benchmarks, especially for detail-intensive tasks (e.g., DocVQA~\cite{mathew2021docvqa}, InfoVQA~\cite{mathew2022infographicvqa}), our method incurs no performance loss. 
This demonstrates that unifying visual inputs into video format and leveraging temporal-based token encoding and compression enables large VLMs to better handle diverse visual tasks.

Our contributions can be summarized as follows:
\begin{itemize}
    \item To address the limitations of separate image and video processing in existing VLMs, we propose \emph{Progressive Visual Token Compression (PVC)} that unifies visual inputs as videos, which allows spatial details and temporal dynamics to be preserved across different modalities.
    \item We introduce a novel \emph{progressive encoding module} and an \emph{adaptive compression module} to progressively encode complement information in visual tokens. This effectively minimizes temporal redundancy while capturing essential spatiotemporal information, allowing a lower token-per-frame in our model.
    \item With the proposed progressive token compression, our model achieves state-of-the-art results across long video and fine-grained short video benchmarks. Meanwhile, it incurs no performance loss on image benchmarks, particularly in detail-sensitive tasks.
\end{itemize}
\section{Related Work}
\label{sec:related}

\noindent\textbf{Vision-Language Models for Images and Videos.}
In the realm of large vision-language models (VLMs), proprietary models~\cite{gemini,gpt4v,gpt4o,sonnet3_5} and open-source models~\cite{chen2024internvl,chen2024far,internvl2,liu2024visual,liu2024improved,li2024llavainterleave,li2024llava,bai2023qwen,wang2024qwen2,zhang2024internlm} have demonstrated exceptional performance across a variety of visual scenarios, including single-image, multi-image, and video contexts.
Large VLMs typically represent an image with hundreds or thousands of visual tokens, but the additional temporal dimension in videos leads to a considerable number of tokens, posing a great challenge.
Some methods~\cite{zhang2024long,liu2024kangaroo,wang2024longllava} extent the context length of the large language model to process more tokens, but this causes a huge computational burden during training and inference.
Other methods uses token compression techniques to reduce the number of tokens~\cite{li2024llava,liu2024oryx,wang2024qwen2}.
For instance, LLaVA-OneVision~\cite{li2024llava} spatially down-samples the video tokens by $2\times2$; 
Oryx MLLM~\cite{liu2024oryx} designs a dynamic compressor for different types of inputs, where videos are down-sampled to a smaller scale;
Qwen2-VL~\cite{wang2024qwen2} leverages pixel shuffle and 3D convs to down-sample vision tokens spatially and temporally, and reduces the resolution of video frames for long videos.

As mentioned above, existing methods usually compress image and video tokens with different strategies to meet the needs of different tasks, such as detailed images and long videos.
Furthermore, the simple spatial or temporal down-sampling does not take full advantage of the temporal redundancy in videos, leading to an inefficient compression.
Our approach, otherwise, unifies the inputs into videos and compresses tokens by exploiting the temporal redundancy.

\vspace{0.3em}
\noindent\textbf{Vision Transformer with Temporal Attention.}
Modern video encoders typically enhance the basic image encoder (e.g., Vision Transformer, ViT
~\cite{dosovitskiy2021an}) with temporal attention to account for the temporal correlations in videos. 
Video Swin Transformer~\cite{liu2022video} applies joint spatio-temporal attention within localized 3D windows. 
TimeSformer~\cite{bertasius2021space}, ViViT~\cite{arnab2021vivit}, TESTA\cite{ren2023testa} and EVLGen~\cite{jian2024expedited} apply self-attention mechanism along the spatial and temporal dimensions, respectively. 
We also introduce an additional temporal attention module to ViT layers, but the attention is causal to realize progressive feature encoding. 
Besides, we replace the Layer Normalization with Adaptive Layer Normalization (AdaLN) injected with timestep conditions to avoid redundant encoding for different input frames.

\vspace{0.3em}
\noindent\textbf{Visual Token Compression.}
Visual token compression is commonly used in VLMs to reduce the number of vision tokens to fit in the context length of LLMs and reduce computational overhead.
Applying pooling~\cite{maaz2024video,zhang2024direct}, down-sampling~\cite{xu2024pllava,li2024llava}, or convolution~\cite{cheng2024videollama} along the spatial and temporal dimensions is straightforward but may cause information loss.
Many other works~\cite{zhang2023video,li2023videochat,li2024mvbench,fei2024video,bai2023qwen,liu2025st,yao2024minicpm} employ Q-Former~\cite{li2023blip} that aggregates visual features into a fixed number of tokens with learnable queries. 
However, learning a cross-attention layer is more difficult, and the compressed tokens may lose spatial and temporal awareness.
Current high-performance VLMs such as InternVL2~\cite{internvl2} and Qwen2-VL~\cite{wang2024qwen2} leverage PixelShuffle that aggregates a local grid of tokens into a single one by concatenating along the feature dimension.
Meanwhile, some works also devise token compression modules specifically for video inputs, for instance, spatio-temporal attentional pooling~\cite{ryoo2024xgen}, and token selection based on the similarity with text queries~\cite{shen2024longvu,li2025llama}.
Nevertheless, these approaches are not applicable to general image and video understanding tasks.

Our progressive compression is based on the high-performance PixelShuffle module. We incorporate AdaLN to extract different spatio-temporal information at different time steps to avoid redundant representation.

\section{Method}

\begin{figure*}[t]
\centering
    \vspace{-1em}
    \includegraphics[width=0.8\linewidth]{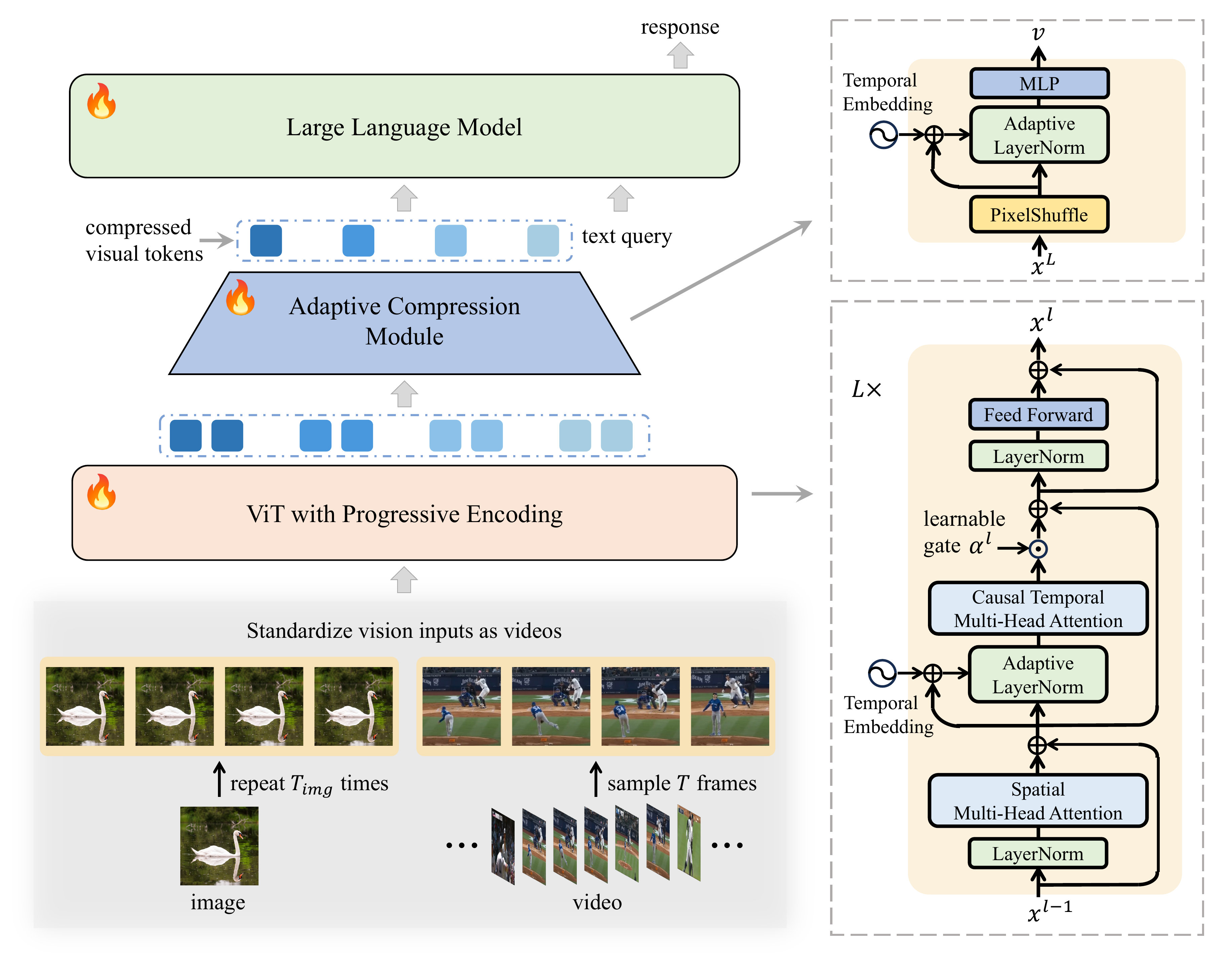}
    \vspace{-0.5em}
    \caption{\textbf{Network architecture of progressive visual token compression (\name{}).} The inputs are standardized as videos, with images repeated to form static videos. A causal temporal attention and an AdaLN layer are incorporated into the ViT layers to progressively encode visual tokens across timesteps. The adaptive compression module, based on PixelShuffle, includes an AdaLN layer to reduce redundancy in visual tokens.}
    \label{fig:overview}
    \vspace{-1.5em}
\end{figure*}

\subsection{Standardize Vision Inputs of VLMs as Videos}

Current high-performance VLMs typically use different methods to compress image and video tokens. Due to limited computational budget, allocating more tokens per image is beneficial for understanding image spatial details, whereas video processing often involves sacrificing the number of tokens per frame to include more frames. 

To address this dilemma, we standardize vision inputs of VLMs as videos, and apply the proposed progressive visual token compression for all videos. Specifically, each image is repeated into multiple frames to create a ``static'' video. Then, progressive compression allows LLMs to revisit the image multiple times and capture richer spatial information in subsequent static frames. This approach enables a unified visual token compression for images and videos, balancing the distinct needs of both modalities. Even when the number of tokens per frame is limited, the necessary spatial details and temporal dynamics can be gradually supplemented in subsequent frame tokens.

For each image $x$ in the inputs, we repeat it $T_{\text{img}}$ times to form a static frame sequence $[x, x, \dots, x]$ (by default $T_{\text{img}}=4$). For each native video input, we uniformly sample $T$ frames, where $T$ is randomly chosen from the range of $[16, 96]$ by default. In scenarios where dynamic resolution~\cite{chen2024far} is used, it is applied to every frame of both static and native videos.

\subsection{Network Architecture}

The network architecture for our proposed \name{} is outlined in Fig.~\ref{fig:overview}, comprising a Vision Transformer (ViT) with Progressive Encoding, an Adaptive Token Compression module, and a Large Language Model (LLM). 
To fully leverage the redundancy between video frames, we integrate temporal attention mechanism into the ViT for progressive encoding. 
Additionally, the adaptive compression module is introduced to effectively reduce the number of visual tokens per frame.

\vspace{0.3em}\noindent\textbf{Vision Transformer with Progressive Encoding.}
To avoid redundant encoding of video frames and give the current frame the potential to only encode new information not extracted from previous frames, we add a temporal attention module within the ViT.
To minimize overhead, temporal attentions are applied only in the last $\tilde{L}$ layers of the ViT (with $L$ layers in total). Let $x$ be the video feature with dimensions [$B$, $T$, $N$, $C$], where $B$ is the batch size, $T$ is the number of frames, $N$ is the number of patches per frame, and $C$ is the hidden dimension. The ViT layer with progressive encoding is defined by:
\begin{align}
	x &\coloneq x + \text{S-MHA}(\text{LayerNorm}(x)), \\
	x &\coloneq x + \alpha \odot \text{T-MHA}(\text{AdaLN}(x;x+\text{TE})), \\
	x &\coloneq x + \text{FFN}(\text{LayerNorm}(x)).
\end{align}

\emph{Spatial Multi-Head Attention (S-MHA)} is the original multi-head self-attention~\cite{vaswani2017attention} in ViT, which is applied on the patch tokens of each frame respectively. 
$x$ is reshaped to dimensions [$B\times T$, $N$, $C$] when input into S-MHA, where $N$ is the sequence length here.

\emph{Temporal Multi-Head Attention (T-MHA)} is an additional \emph{causal} multi-head self-attention~\cite{vaswani2017attention} applied on the temporal dimension of the video feature.
Each patch token attends to the tokens at the same spatial position in the previous frames.
Specifically, x is rearranged to dimensions [$B\times N$, $T$, $C$] when input into T-MHA, where $T$ is the sequence length here.

\emph{Temporal Embedding (TE)} is added to encode temporal position information. First, we use relative timestamps to indicate the video frames: $t=\left[0,\frac{1}{T-1},\dots,\frac{T-2}{T-1}, 1\right]$.
Then we encode all timestamps by a 256-dimension sinusoidal position embeddings~\cite{vaswani2017attention}, denoted as $\tilde{t}\in\mathbb{R}^{T\times256}$.
After that, we obtain the temporal embedding via an MLP:
\begin{equation}
    \text{TE} = W_2\cdot\text{SiLU}\left(W_1\tilde{t}\right)
\end{equation}
where $W_1$ and $W_2$ are learnable parameters, SiLU refers to Sigmoid Linear Unit~\cite{hendrycks2016gaussian}.
 
\emph{Adaptive Layer Normalization (AdaLN)} was first proposed by \cite{peebles2023scalable} and is now widely used to adjust the normalization parameters according to given conditions, making the model more adaptable to different task requirements. In order to adapt to the ``static'' videos where different frames are the same, we use AdaLN here so that the network can extract different spatial information at different time steps and avoid redundant encoding.
$\text{AdaLN}(x;z)$ normalizes the input $x$ and applies an affine operation with the coefficients conditioned on $z$:
\begin{equation}
    \text{AdaLN}(x;z) = \gamma(z)\odot\text{LayerNorm}(x) + \beta(z).
\end{equation}
The affine scale $\gamma(z)$ and bias $\beta(z)$ are obtained via:
\begin{equation}
    \gamma(z) = W_4\cdot\text{SiLU}(W_3z), ~ \beta(z) = W_6\cdot\text{SiLU}(W_5z),
\end{equation}
where $W_3$, $W_4$, $W_5$ and $W_6$ are learnable parameters.

Additionally, a \emph{learnable gate} $\alpha \in \mathbb{R}^C$ is introduced to scale the output of T-MHA.
It is zero-initialized to make sure that the ViT is initialized as the pre-trained checkpoint.

\vspace{0.3em}\noindent\textbf{Adaptive Compression.}
To reduce the computational budget, current VLMs~\cite{chen2024far,wang2024qwen2,li2024llava} commonly shorten the sequence length by compressing visual tokens. For instance, InternVL-1.5~\cite{chen2024far} employs a PixelShuffle~\cite{shi2016real} operation~\footnote{Originally from the super-pixel area~\cite{shi2016real}; here, it refers to concatenating 4 tokens from each adjacent 2$\times$2 region along the channel dimension to form a single token.} followed by an MLP, reducing the visual token length by a factor of 4. However, this relies on a shared MLP to compress different visual sequences, which hinders reducing the representation redundancy between different frames.

To this end, we introduce an adaptive compression module by integrating an AdaLN~\cite{peebles2023scalable} layer before the shared MLP. The reason for using AdaLN here is the same as in the progressive encoding module, allowing the network to extract different spatiotemporal information at different time steps to avoid redundant representation.
The input of token compression is the output video tokens $x$ with dimensions [$B$, $T$, $N$, $C$] from the ViT.
It compresses the number of tokens per frame from $N$ to $M$ as follows:
\begin{equation}
	\tilde{x} = \text{PixelShuffle}(x)\in\mathbb{R}^{B\times T\times M\times (\frac{N}{M} \cdot C)}
\end{equation}
\begin{equation}
	v = \text{MLP}(\text{AdaLN}(\tilde{x};\tilde{x}+\text{TE})).
\end{equation}
where AdaLN and TE is the same as mentioned above.

By default, we set $M=\frac{1}{16}N$. Correspondingly, the kernel size for PixelShuffle is set to $4\times4$, \ie, it aggregates a $4\times4$ grid of tokens into a single token by concatenating along the feature dimension. Note that the compression ratio of 16 is higher than that in previous approaches (typically set to 4). For image modality, we repeat each image $T_{\text{img}}$ times ($T_{\text{img}}=4$ by default), resulting in the same visual token length as previous approaches for image input.

\subsection{Training Strategies and Data}

\noindent\textbf{Training Strategies.}
Following common practices from previous VLMs~\cite{liu2024visual,li2024llava,chen2024far,internvl2,bai2023qwen,wang2024qwen2}, the training of \name{} consists of two stages: the pre-training stage and the instruction-tuning stage. 
The main difference is that all parameters of \name{} are jointly trained in both stages, without freezing ViT or LLM as in previous approaches~\cite{liu2024visual,li2024llava,chen2024far,internvl2,bai2023qwen,wang2024qwen2}. 
We empirically find that this strategy works better for \name{} (see Appendix for ablation experiments). Additionally, in the pre-training stage, we use a larger learning rate (10$\times$) for randomly initialized parameters compared to pre-trained parameters.

\vspace{0.3em}\noindent\textbf{Pre-training Data.}
The pre-training data used in \name{} comprises two parts: (1) 45M image-text pairs used in InternVL2's~\cite{internvl2} pre-training stage, including LAION~\cite{schuhmann2022laion}, COYO~\cite{kakaobrain2022coyo-700m}, \etc. 
(2) 23M video-text pairs, including InternVid~\cite{wang2023internvid}, WebVid~\cite{bain2021frozen}, \etc.
The sampling ratio of image-text to video-text data is set to 1:1.

\vspace{0.3em}\noindent\textbf{Instruction Tuning Data.}
The instruction tuning data used in \name{} consists of two parts:
(1) 7.2M image-text and video-text data used in InternVL2's~\cite{internvl2} fine-tuning stage.
(2) 3.4M video-text data collected and refined from Vript~\cite{yang2024vript}, LLaVA-Video~\cite{zhang2024video}, \etc.
The sampling ratio of image-text to video-text data is set to 3:1.

For further details of training data, please refer to the Appendix.
Note that all the image-text data used in \name{} are adopted from InternVL2~\cite{internvl2}, and we conduct experiments to assess the effect of the newly appended video-text data (see Appendix).
\section{Experiment}

\setlength{\tabcolsep}{3pt}
\setlength{\doublerulesep}{2\arrayrulewidth}
\renewcommand{\arraystretch}{1.1}

\begin{table*}[t]
\centering
\small
\caption{\textbf{Video-language benchmark results}. Our proposed PVC is compared with image-video general VLMs and video VLMs. ``LongVideo.'' refers to LongVideoBench. The results of VideoMME are reported as ``without subscript/with subscript''. The best results among image-video general VLMs are \textbf{bold}, and the best results among all open-source VLMs are \underline{underlined}. * Native resolution is used without fixed \# token/frame. $\dagger$ 1024 tokens in total, with 16 or 96 frames used for different tasks.}
\vspace{-0.5em}
\begin{tabular}{lcc|cccccccc}
\toprule
Model & Size & \makecell{\# token\\/frame}& \rotatebox{45}{MVBench} & \rotatebox{45}{VideoMME} & \rotatebox{45}{MLVU} & \rotatebox{45}{LongVideo.} & \rotatebox{45}{NextQA} & \rotatebox{45}{Egoschema} & \rotatebox{45}{PercepTest} & \rotatebox{45}{ActNet-QA} \\

\midrule
\midrule
\multicolumn{4}{l}{\emph{\demph{Commercial VLMs:}}} \\
\midrule
\demph{Gemini-1.5-Pro~\cite{gemini}} & \demph{-} & \demph{-} & \demph{52.6} & \demph{75.0/81.3} & \demph{-} & \demph{64.0} & \demph{-} & \demph{64.5} & \demph{-} & \demph{57.5} \\
\demph{GPT-4V~\cite{gpt4v}} & \demph{-} & \demph{-} & \demph{43.7} & \demph{59.9/63.3} & \demph{49.2} & \demph{61.3} & \demph{-} & \demph{55.6} & \demph{-} & \demph{57.0} \\
\demph{GPT-4o~\cite{gpt4o}} & \demph{-} & \demph{-} & \demph{47.8} & \demph{71.9/77.2} & \demph{64.6} & \demph{66.7} & \demph{-} & \demph{72.2} & \demph{-} & \demph{-} \\
\midrule 
\midrule
\multicolumn{4}{l}{\emph{Video VLMs:}} \\
\midrule
Video-CCAM~\cite{fei2024video} & 4B & 64/10.7$^\dagger$ & 62.8 & 50.1/51.2 & 56.5 & - & - & - & - & 58.0 \\
BLIP-3-Video~\cite{ryoo2024xgen} & 4B & 16 & - & - & - & - & 77.1 & - & - & 56.9 \\
\midrule
VideoChat2~\cite{li2024mvbench} & 7B & 96 & 60.4 & 39.5/43.8 & 47.9 & - & - & 54.4 & - & 49.1 \\ 
PLLaVA~\cite{xu2024pllava} & 7B & 144 & 46.6 & - & - & - & - & 54.5 & - & 56.3 \\
VideoLLaMA2~\cite{cheng2024videollama} & 7B & 72 & 54.6 & 47.9/50.3 & 48.5 & - & - & 51.7 & 51.4 & 50.2 \\ 
Tarsier~\cite{wang2024tarsier} & 7B & 576 & 62.6 & - & - & - & 71.6 & 49.9 & 59.5 \\
LongVU~\cite{shen2024longvu} & 7B & 64-144 & 66.9 & - /60.6 & 65.4 & - & - & \underline{67.6} & - & - \\
LLaVA-Video~\cite{zhang2024video} & 7B & 169 & 58.6 & 63.3/\underline{69.7} & 70.8 & 58.2 & \underline{83.2} & 57.3 & 67.9 & 56.5 \\ 
Kangaroo~\cite{liu2024kangaroo} & 8B & 1024 & 61.1 & 56.0/57.6 & 61.0 & 54.8 & - & 62.7 & - & - \\
InternVideo2~\cite{wang2024internvideo2} & 8B & 96 & 67.2 & - & - & - & - & 60.0 & 63.4 & - \\
 Video-CCAM~\cite{fei2024video} & 9B & 64/10.7$^\dagger$ & 64.6 & 50.3/52.6 & 58.5 & - & - & - & - & \underline{59.7} \\
\midrule 
\midrule
\multicolumn{4}{l}{\emph{Image-video general VLMs:}} \\
\midrule
Qwen2-VL~\cite{wang2024qwen2} & 2B & -* & 63.2 & 55.6/60.4 & - & - & - & 54.9 & 53.9 & - \\
InternVL2~\cite{internvl2} & 2B & 256 & 60.2 & 45.0/47.3 & 50.4 & - & - & 46.7 & 49.9 & -  \\
\rowcolor{Gray!15}
\textbf{PVC$_{\text{InternVL2}}$ (Ours)} & 2B & 64 & 69.4 & 54.5/56.7 & 63.4 & 51.6 & 73.4 & 49.8 & 62.9 & 52.8 \\
\midrule
LongVA~\cite{zhang2024long} & 7B & 144 & - & 52.6/54.3 & 56.3 & - & 68.3 & - & - & 50.0 \\
IXC-2.5~\cite{zhang2024internlm} & 7B & 400 & 69.1 & 55.8/58.8 & 58.8 & - & 71.0 & - & 34.4 & 52.8 \\
LLaVA-OV~\cite{li2024llava} & 7B & 196 & 56.7 & 58.2/61.5 & 64.7 & 56.5 & 79.4 & 60.1 & 57.1 & 56.6 \\
Oryx MLLM~\cite{liu2024oryx} & 7B & -* & 63.9 & 58.3/62.6 & 67.5 & 55.3 & 81.9 & - & \underline{\textbf{68.6}} & - \\
Qwen2-VL~\cite{wang2024qwen2} & 7B & -* & 67.0 & 63.3/69.0 & - & - & - & \textbf{66.7} & 62.3 & - \\
InternVL2~\cite{internvl2} & 8B & 256 & 66.4 & 54.0/56.9 & 52.0 & - & - & 55.0 & 52.0 & - \\
\rowcolor{Gray!15}
\textbf{PVC$_{\text{InternVL2}}$ (Ours)} & 8B & 64 & \underline{\textbf{73.8}} & \underline{\textbf{64.1}}/\underline{\textbf{69.7}} & \underline{\textbf{72.4}} & \underline{\textbf{59.2}} & \textbf{82.0} & 59.6 & 68.4 & \textbf{57.1} \\
\bottomrule
\end{tabular}
\vspace{-1em}
\label{tab:video_benchmark}
\end{table*}

\subsection{Experiment Settings}

\noindent\textbf{Implementation Details.}
Our implementation mainly follows the InternVL series~\cite{chen2024far,internvl2}, referred to as \name{}$_\text{InternVL2}$ below.
We adopt ViT-L/14~\cite{dosovitskiy2021an} as the visual encoder, initializing its weights from InternViT-300M-448px-V1.5~\cite{chen2024far}. The progressive encoding is introduced in the last 8 layers of the 24-layer ViT (\ie, $\tilde{L}=8$ and $L=24$), whose T-MHA is randomly initialized. InternLM2-Chat-1.8B~\cite{cai2024internlm2} is employed as the LLM for \name{}$_\text{InternVL2}$-2B, while InternLM2.5-Chat-7B~\cite{cai2024internlm2} is utilized for \name{}$_\text{InternVL2}$-8B. We use the Adam optimizer~\cite{kingma2014adam} for both the pre-training and instruction-tuning stages, setting the learning rate to 2e-4 during pre-training and 4e-5 for fine-tuning. In the pre-training stage, the learning rate for weights initialized from pre-trained ViT and LLM models is reduced by a factor of 10 (\ie, 2e-5). Weight decay is set to 0.01 in both stages. Our model is pre-trained on 50M samples and fine-tuned for 1 epoch. The input resolution for the visual encoder is set to 448$\times$448, with dynamic resolution~\cite{chen2024far} enabled for image data in both stages. The number of image tiles ranges up to 12, based on the image's aspect ratio and resolution.

\vspace{0.3em}\noindent\textbf{Evaluation Benchmarks.}
To validate \name{}'s general capabilities, we evaluate it on various image-language and video-language benchmarks. The image-language benchmarks include AI2D~\cite{kembhavi2016diagram}, ChartQA~\cite{masry2022chartqa}, DocVQA~\cite{mathew2021docvqa}, InfoVQA~\cite{mathew2022infographicvqa}, SQA~\cite{lu2022learn}, TextVQA~\cite{singh2019towards}, MMB~\cite{liu2025mmbench}, MME~\cite{fu2024mme}, MMMU~\cite{yue2024mmmu}, SEED-I~\cite{li2023seed}, and OCRBench~\cite{liu2023hidden}. For video-language benchmarks, we use MVBench~\cite{li2024mvbench}, VideoMME~\cite{fu2024video}, MLVU~\cite{zhou2024mlvu}, LongVideoBench~\cite{wu2024longvideobench}, NextQA~\cite{xiao2021next}, Egoschema~\cite{mangalam2023egoschema}, PercepTest~\cite{patraucean2024perception}, and ActNet-QA~\cite{yu2019activitynet}.

\subsection{Comparison with State-of-the-art Methods}
In this section, we evaluate our \name{} across a wide range of multimodal benchmarks to validate its understanding and reasoning abilities. 
We compare our \name{} with state-of-the-art VLMs within 8B parameters, including commercial VLMs~\cite{gpt4o,gpt4v,gemini}, open-source general VLMs~\cite{internvl2,wang2024qwen2,zhang2024long,zhang2024internlm,li2024llava,liu2024oryx}, and open-source video VLMs~\cite{fei2024video,ryoo2024xgen,li2024mvbench,xu2024pllava,cheng2024videollama,wang2024tarsier,shen2024longvu,zhang2024video,liu2024kangaroo,wang2024internvideo2}. 
As shown in Tab.~\ref{tab:video_benchmark} and Tab.~\ref{tab:image_benchmark}, \name{} achieves superior performance on long video tasks (\eg, VideoMME, MLVU) and fine-grained video tasks (\eg, MVBench), while maintaining very competitive performance on image tasks.

\setlength{\tabcolsep}{2pt}
\setlength{\doublerulesep}{2\arrayrulewidth}
\renewcommand{\arraystretch}{1.1}

\begin{table*}[t]
\centering
\small
\caption{\textbf{Image-language benchmark results}. We compare our PVC with VLMs that support both image and video inputs. * Native resolution is used without fixed \# token/image tile.}
\vspace{-0.5em}
\resizebox{1.0\linewidth}{!}{
\begin{tabular}{lcc|ccccccccccc}
\toprule
\multirow{2}{*}{Model} & \multirow{2}{*}{Size} & \multirow{2}{*}{\makecell{\# token\\/image tile}} & AI2D & ChartQA & DocVQA & InfoVQA & SQA & TextVQA & MMB & MME & MMMU & SEED-I & OCRBench \\ \cline{4-14} 
& & & test & test & test  & test & test & val & en-test & sum & val & - & - \\
\midrule
\midrule
\multicolumn{4}{l}{\emph{\demph{Commercial VLMs:}}} \\
\midrule
\demph{Gemini-1.5-Pro~\cite{gemini}} & \demph{-} & \demph{-}& \demph{94.4} & \demph{87.2} & \demph{93.1} & \demph{81.0} & \demph{-} & \demph{-} & \demph{-} & \demph{-} & \demph{62.2} & \demph{-} & \demph{-} \\
\demph{GPT-4V~\cite{gpt4v}} & \demph{-} & \demph{-} & \demph{78.2} & \demph{78.5} & \demph{88.4} & \demph{-} & \demph{75.7} & \demph{-} & \demph{-} & \demph{1926} & \demph{56.8} & \demph{49.9} & \demph{-} \\
\demph{GPT-4o~\cite{gpt4o}} & \demph{-} & \demph{-} & \demph{94.2} & \demph{85.7} & \demph{92.8} & \demph{-} & \demph{-} & \demph{-} & \demph{83.4} & \demph{2329} & \demph{69.1} & \demph{76.2} & \demph{736} \\
\midrule 
\midrule
\multicolumn{4}{l}{\emph{Image-video general VLMs:}} \\
\midrule
Qwen2-VL~\cite{wang2024qwen2} & 2B & -* & 74.7 & 73.5 & 90.1 & 65.5 & - & 79.7 & 74.9 & 1872 & 41.1 & - & 809 \\
InternVL2~\cite{internvl2} & 2B & 256 & 74.1 & 76.2 & 86.9 & 58.9 & 94.0 & 73.4 & 73.2 & 1877 & 34.3 & 71.6 & 784 \\
\rowcolor{Gray!15}
\textbf{PVC$_\text{InternVL2}$ (Ours)} & 2B & 64 & 76.3 & 78.9 & 87.7 & 59.9 & 94.9 & 76.2 & 75.8 & 1960 & 36.9 & 73.2 & 796 \\
\midrule
LongVA~\cite{zhang2024long} & 7B & 144 & 70.7 & 70.4 & 80.8 & 49.4 & - & - & - & - & 42.6 & - & - \\
IXC-2.5~\cite{zhang2024internlm} & 7B & 400 & 81.5 & 82.2 & 90.9 & 69.9 & - & 78.2 & 82.2 & 2229 & 42.9 & 75.4 & 690 \\
LLaVA-OV~\cite{li2024llava} & 7B & 729 & 81.4 & 80.0 & 87.5 & 68.8 & 96.0 & - & - & 1998 & 48.8 & 75.4 & - \\
Oryx MLLM~\cite{liu2024oryx} & 7B & -* & 78.5 & - & 89.0 & - & - & 75.0 & 81.4 & - & 43.9 & - & 672 \\
Qwen2-VL~\cite{wang2024qwen2} & 7B & -* & 83.0 & 83.0 & \textbf{94.5} & \textbf{76.5} & - & \textbf{84.3} & 83.0 & \textbf{2327} & \textbf{54.1} & - & \textbf{866} \\
InternVL2~\cite{internvl2} & 8B & 256 & \textbf{83.8} & 83.3 & 91.6 & 74.8 & 97.1 & 77.4 & 81.7 & 2210 & 49.3 & 76.2 & 794 \\
\rowcolor{Gray!15}
\textbf{PVC$_\text{InternVL2}$ (Ours)} & 8B & 64 & \textbf{83.8} & \textbf{84.1} & 92.5 & 75.0 & \textbf{97.7} & 80.0 & \textbf{83.9} & 2282 & 50.9 & \textbf{77.2} & 807 \\
\bottomrule
\end{tabular}}
\vspace{-1em}
\label{tab:image_benchmark}
\end{table*}

\vspace{0.3em}\noindent\textbf{Video-Language Benchmarks.}
The results on video-language benchmarks are presented in Tab.~\ref{tab:video_benchmark}, which can be summarized as the following:

\noindent\textbf{(1)} Our \name{}$_{\text{InternVL2}}$ utilizes 64 tokens per frame, which is fewer than most existing methods (\eg, LLaVA-Video~\cite{zhang2024video} uses 169 tokens, InternVL2~\cite{internvl2} uses 256 tokens). However, \name{}$_{\text{InternVL2}}$ achieves strong performance on all video tasks. In contrast, previous approaches such as LongVU~\cite{shen2024longvu} and VideoCCAM~\cite{fei2024video} also reduce tokens per frame to 64 but exhibit lower performance on the fine-grained video understanding task, MVBench. It demonstrates that our \name{} can progressively capture the information from video frames and effectively exploit the temporal redundancy (see the ablation study in Tab~\ref{tab:ablation} for evidence.)

\noindent\textbf{(2)} Our \name{}$_{\text{InternVL2}}$-8B attains superior performance on MVBench, VideoMME, MLVU, and LongVideoBench, demonstrating its strong capability in both fine-grained and long video understanding. Notably, \name{}$_{\text{InternVL2}}$-8B reaches 73.8 on MVBench, significantly surpassing the previous best accuracy of 69.1 among open-source models within 8B parameters, set by IXC-2.5~\cite{zhang2024internlm}.

\noindent\textbf{(3)}
Our \name{}$_{\text{InternVL2}}$-8B also obtains the best results in NextQA and ActNet-QA among image-video general VLMs.
On Egoschema and PercepTest, our model is inferior to some existing models, which may be caused by the lack of training data in some specific domains.

\vspace{0.3em}\noindent\textbf{Image-Language Benchmarks.}
The results on image-language benchmarks are presented in Tab.~\ref{tab:image_benchmark} and can be summarized as follows:

\noindent\textbf{(1)} \name{}$_{\text{InternVL2}}$ uses the same architecture and image-text training data as InternVL2~\cite{internvl2}. In comparison with InternVL2, \name{}$_{\text{InternVL2}}$ does not compromise performance on image tasks, even on OCR-related tasks (\eg, DocVQA, InfoVQA, \etc) that heavily depend on spatial details (as shown in Tab.~\ref{tab:ablation}, directly reducing the tokens per patch would significantly hurt performance on DocVQA and InfoVQA).
This validates that \name{} can progressively encode spatial information from repeated images through the temporal attention module, and effectively compress the visual tokens to fully preserve spatial details.
Notably, \name{}$_{\text{InternVL2}}$-8B improves TextVQA performance from 77.4 to 80.0 (+2.6). 

\noindent\textbf{(2)} Compared to Qwen2-VL-7B~\cite{wang2024qwen2}, our \name{}$_{\text{InternVL2}}$-8B achieves comparable or better performance on AI2D (+0.8), ChartQA (+1.1), and MMB (+0.9), but lags behind on DocVQA (-2.0), InfoVQA (-1.5), TextVQA (-4.3), MME (-45), MMMU (-3.2), and OCRBench (-59). The benchmarks where Qwen2-VL excels are primarily OCR-related (4 out of 6), which are highly sensitive to OCR-specific training data. Given the differences in model architecture and training data, a direct comparison is challenging. 
However, our \name{} is orthogonal and complementary to the advances in model architecture and training data. The ablation studies in Tab.~\ref{tab:ablation} show the effectiveness of our \name{}.

\subsection{Ablation Study}
\setlength{\tabcolsep}{3pt}
\setlength{\doublerulesep}{2\arrayrulewidth}
\renewcommand{\arraystretch}{1.1}

\begin{table*}[t]
\centering
\small
\caption{\textbf{Ablation of key components in PVC.} The ablation study is conducted on 2B model with shortened pre-training on 10M samples. ``Adapt. compress'' refers to adaptive compression. VideoMME is tested under ``without subscript'' setting.}
\vspace{-0.5em}
\begin{tabular}{c|ccccc|ccc|ccc}
\Xhline{2\arrayrulewidth}
Settings & \makecell{\# tokens\\per image} & \makecell{\# image\\repeat} & \makecell{\# video\\frames} & \makecell{Progressive\\encoding} & \makecell{Adapt. \\compress} & MVBench & VideoMME & MLVU & DocVQA & InfoVQA & MMB \\
\hline
(a) & 256 & 1 & 16 & & & 61.9 & 44.9 & 50.5 & 86.4 & 59.1 & 75.1 \\
(b) & 64 & 4 & 64 & & & 58.0 & 46.0 & 53.0 & 82.7 & 52.6 & 73.1 \\
(c) & 64 & 4 & 64 & $\checkmark$ & & 61.8 & 46.5 & 54.2 & 85.9 & 58.1 & 74.8 \\
\name{} & 64 & 4 & 64 & $\checkmark$ & $\checkmark$ & 62.4 & 46.7 & 55.1 & 86.7 & 58.5 & 74.9 \\
\Xhline{2\arrayrulewidth}
\end{tabular}
\vspace{-0.5em}
\label{tab:ablation}
\end{table*}

\begin{figure*}[t]
  \centering
    \subfloat{
        \includegraphics[width=0.22\linewidth]{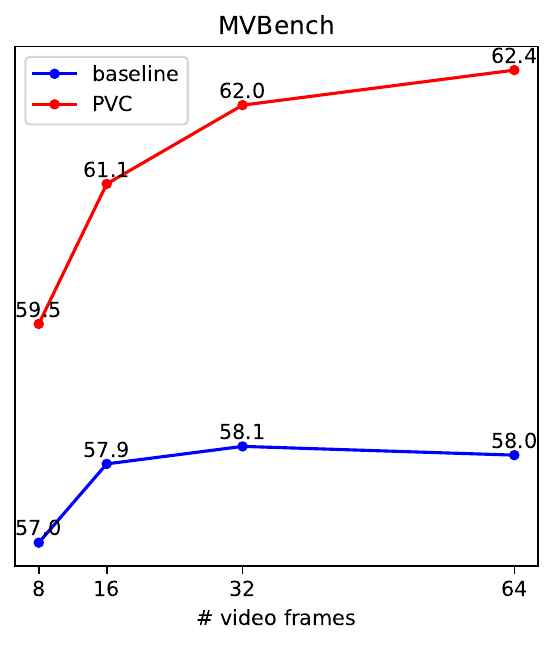}
}
    \subfloat{
        \includegraphics[width=0.22\linewidth]{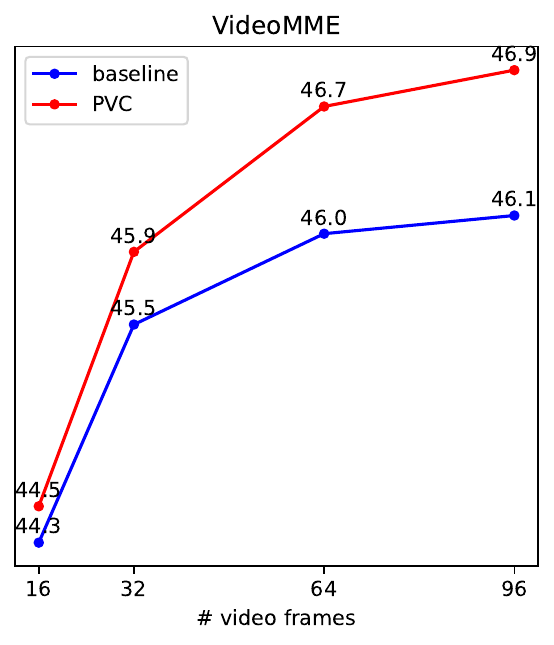}
}
    \subfloat{
        \includegraphics[width=0.22\linewidth]{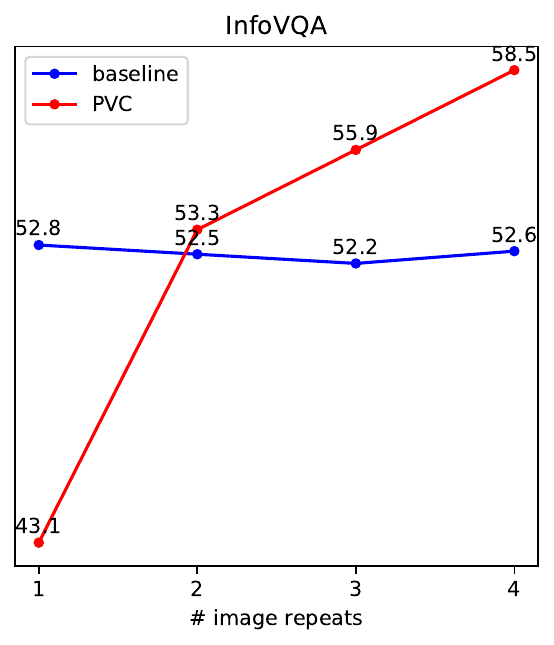}
}
    \subfloat{
        \includegraphics[width=0.22\linewidth]{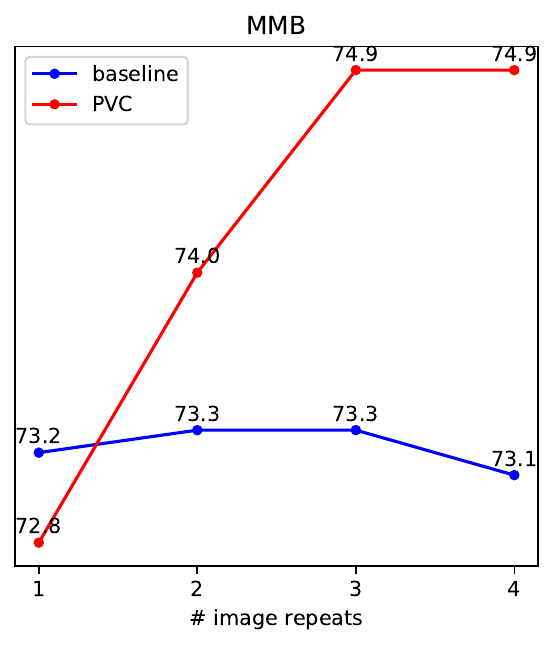}
}
\vspace{-0.5em}
\caption{\textbf{Analysis of progressive compression.} We compare our PVC model with the baseline without progressive compression (setting (b)) in Tab.~\ref{tab:ablation}. For video tasks (MVBench and VideoMME), we test different number of input frames. For image tasks (InforVQA and MMB), we test different repetition times of the image.}
\vspace{-1.5em}
\label{fig:analysis}
\end{figure*}

In this section, we present a series of ablation experiments to validate the effectiveness of \name{}. To reduce experimental burden, the ablation study is conducted on 2B model with shortened pre-training on 10M samples. Different ablation settings are evaluated on representative video-language benchmarks (MVBench, VideoMME, and MLVU) and image-language benchmarks (DocVQA, InfoVQA, and MMB). The results are presented in Tab.~\ref{tab:ablation}.

\vspace{0.2em}\noindent\textbf{Standardize Vision Inputs as Videos.}
The standardization reduces tokens per image from 256 to 64 while maintaining the overall number of visual tokens by repeating each image and increasing the number of video frames. As shown in settings (a) and (b) of Tab.~\ref{tab:ablation}, directly standardizing vision inputs as video significantly reduces performance on OCR-related image tasks (DocVQA -3.7 and InfoVQA -6.5) and fine-grained video tasks (MVBench -3.9). This performance drop occurs because repeating the image four times for the vanilla ViT and vision compression module only produces four groups of identical visual tokens, with repeated information encoded. Consequently, spatial details are lost compared to using 256 distinct visual tokens. However, long-video understanding tasks (VideoMME +1.1 and MLVU +2.5), which do not require as much visual detail, can benefit from the increase in video frames.

\vspace{0.3em}\noindent\textbf{Progressive Encoding.}
Progressive encoding involves adding temporal causal attention in ViT. As shown in Tab.~\ref{tab:ablation}, settings (b) and (c) demonstrate that adding temporal attention in ViT mitigates the information loss observed in setting (b) and maintains comparable performance with setting (a) on tasks requiring spatial details (MVBench, DocVQA, and InfoVQA). Furthermore, temporal attention significantly boosts performance on long video tasks (MLVU +1.2 and VideoMME +0.5). These results validate that our proposed progressive encoding techniques can effectively encode both spatial details and temporal correlations.

\vspace{0.3em}\noindent\textbf{Adaptive Compression.}
As shown in Tab.~\ref{tab:ablation} setting (c) and PVC, the performance on all tasks is further improved by adding the adaptive compression module, especially on long video tasks (\eg, MLVU +0.9). This demonstrates our proposed adaptive compression mechanism can better exploit representation redundancy within video frames and model long-term temporal correlations.

\subsection{Analysis}
\noindent\textbf{Effectiveness of Progressive Compression.} We evaluate our model by gradually increasing the number of video frames and image repetitions.
We use the model in Tab.~\ref{tab:ablation} setting (b) (without progressive compression) as the baseline.
The results are shown in Fig.~\ref{fig:analysis}.

For MVBench, which mainly contains videos shorter than one minute with fine-grained queries, the baseline model's score plateaus after increasing to 16 frames, while our model still gets better with complement information encoded.
For the long-video task VideoMME, the scores of both models increase with the number of frames. Still, our model grows faster, indicating that it is better at removing temporal redundancy and capturing temporal dynamics.

On image tasks, increasing the number of image repetitions does not boost the baseline model's performance since its repeated visual tokens do not provide more information.
On the detail-sensitive task InfoVQA, our model performs poorly when the image is repeated only once because only part of the information is extracted.
When the image is repeated more times, the detailed information is supplemented, so the results get substantially better.
For MMB, the images contain less spatial details, so the gain from repeating more times is relatively small for our model.

\setlength{\tabcolsep}{3pt}
\setlength{\doublerulesep}{2\arrayrulewidth}
\renewcommand{\arraystretch}{1.1}

\begin{table}[t]
\centering
\small
\caption{\textbf{Speed of image repetition and progressive encoding.} We report the FLOPs and FPS (samples per second) for processing samples containing a 448$\times$448 image and 2048 text tokens, measured on an NVIDIA A100.}
\vspace{-0.5em}
\resizebox{1.0\linewidth}{!}{
\begin{tabular}{l|ccc|cc}
\Xhline{2\arrayrulewidth}
Model & \makecell{\# tokens\\per image} & \makecell{\# image\\repeats} & \makecell{Progressive\\encoding} & FLOPs & FPS \\
\hline
InternVL2-8B & 256 & 1 &  & 13.3T & 4.8\\
PVC$_{\text{InternVL2}}$-8B & 64 & 4 & $\checkmark$ & 14.1T & 4.5 \\
\hline
\makecell[c]{$\Delta_{\text{relative}}$} & - & - & - & +6.0\% & -6.3\% \\
\Xhline{2\arrayrulewidth}
\end{tabular}
}
\vspace{-1.5em}
\label{tab:image_repeat}
\end{table}

\vspace{0.3em}\noindent\textbf{Speed of Image Repetition and Progressive Encoding.}
As shown in Tab.~\ref{tab:image_repeat}, for the 8B model, repeating images introduces only a marginal overhead (+6.0\% FLOPs relative, -6.3\% FPS relative), since the visual feature computation can be partially reused and the main computational load is in the LLM. Thus, the inference speed of our \name{}$_{\text{InternVL2}}$ on image tasks remains similar to that of InternVL2.

\vspace{-0.3em}
\section{Conclusion}
\vspace{-0.3em}

In conclusion, our work introduces Progressive Visual Token Compression (PVC), a unified approach for effectively processing images and videos within large Vision-Language Models (VLMs). This strategy leverages the standardization of visual inputs as videos, alongside a progressive encoding module and an adaptive compression module. Extensive empirical analysis demonstrates that PVC can gradually capture both spatial details and temporal dynamics from the sequence of image repetitions and video frames. Consequently, PVC not only unifies image and video processing, but also achieves state-of-the-art performance on fine-grained short video tasks and long video tasks.
It also maintains accuracy on image benchmarks, particularly in detail-sensitive tasks.

\vspace{0.5em}
\noindent\textbf{Acknowledgments.} This work is supported by the National Key R\&D Program of China (NO. 2022ZD0161300), by the National Natural Science Foundation of China (62376134).

\section*{Appendix}
\appendix

\section{Training Details}
Hyper-parameters used in the pre-training and instruction tuning of PVC are listed in Tab.~\ref{tab:hyper}.
\setlength{\tabcolsep}{2pt}
\setlength{\doublerulesep}{2\arrayrulewidth}
\renewcommand{\arraystretch}{1.1}

\begin{table}[h]
    \renewcommand{\thetable}{5}
    \centering
    \small
    \caption{\textbf{Training hyper-parameters of PVC.}}
    \label{tab:hyper}
    \resizebox{1.0\linewidth}{!}{
    \begin{tabular}{lcc}
        \toprule
        Training stage & Pre-training & Instruction tuning \\
        \midrule
        Max sequence length & \multicolumn{2}{c}{8192} \\
        Max tile/image & \multicolumn{2}{c}{12} \\
        Token/frame (tile) & \multicolumn{2}{c}{64} \\
        Number of image repeats & \multicolumn{2}{c}{4} \\
        Number of video frames & \multicolumn{2}{c}{16-96, uniform} \\
        Optimizer &   \multicolumn{2}{c}{AdamW} \\
        Learning rate & 2e-4 & 4e-5 \\
        Weight decay & \multicolumn{2}{c}{0.01 (2B), 0.05 (8B)}\\
        Optimizer momentum & \multicolumn{2}{c}{$\beta_1, \beta_2=0.9, 0.999$} \\
        Learning rate schedule & constant with warmup & cosine decay\\
        Warmup steps & 100 & 240 \\
        Training steps & 25k & 8k \\
        Batch size & 2048 & 1024 \\
        \bottomrule
    \end{tabular}
    }
    \vspace{-1em}
\end{table}

\section{Dataset Details}
\begin{table}[ht]\scriptsize
\renewcommand{\arraystretch}{1.2}
\renewcommand{\thetable}{6}
\caption{\textbf{Summary of datasets used in the pre-training stage.} *IT refers to the instruction tuning data in VideoChat2.
}

\begin{subtable}{0.47\textwidth}
    \setlength\tabcolsep{6.4pt}
    \centering 
    \begin{tabular}{ll}
        \multicolumn{1}{l|}{task} & dataset \\
        \hline
        \multicolumn{1}{l|}{Short Caption} & \makecell[l]{Laion (en\&zh)~\cite{schuhmann2022laion}, COYO~\cite{kakaobrain2022coyo-700m}, COCO~\cite{lin2014microsoft}} \\
        
        \rowcolor{gray!15} \multicolumn{1}{l|}{OCR} & Wukong-OCR~\cite{gu2022wukong}, LaionCOCO-OCR~\cite{schuhmann2022laion-coco} \\
        \multicolumn{1}{l|}{Detection} & GRIT~\cite{peng2023kosmos}, Objects365~\cite{shao2019objects365} \\
        
        \rowcolor{gray!15} \multicolumn{1}{l|}{Conversation} & All-Seeing (en\&zh)~\cite{wang2023all} \\ 

        \multicolumn{2}{l}{Image-text instruction data (see Tab.~\ref{tab:ds_sft_image})} \\

    \end{tabular}
    \caption{Image-text datasets used in the pre-training stage.}
\label{tab:ds_pretrain_image}
\end{subtable}

\begin{subtable}{0.47\textwidth}
    \setlength\tabcolsep{6.4pt}
    \begin{tabular}{l|l}
          task & dataset \\
\hline
                                        & InternVid-10M~\cite{wang2023internvid}, WebVid~\cite{bain2021frozen}, OpenVid~\cite{nan2024openvid},    \\
\multirow{-2}{*}{Short Caption}         & TextVR~\cite{wu2025large}    \\

\rowcolor{gray!15}
                                        & MiraData~\cite{ju2024miradata}, ShareGPT4Video~\cite{chen2024sharegpt4video}, Vript~\cite{yang2024vript},            \\
\rowcolor{gray!15}
\multirow{-2}{*}{Detailed Caption}      & Vript\_Chinese~\cite{yang2024vript}, LSMDC~\cite{rohrbach2015dataset}  \\

                                        & STAR~\cite{wu2024star}, VideoGPT+~\cite{maaz2024videogpt+}, EgoTaskQA~\cite{jia2022egotaskqa}     \\
\multirow{-2}{*}{VQA}                   & CLEVRER~\cite{yi2019clevrer}, Mementos~\cite{wang2024mementos}     \\

\rowcolor{gray!15}
Classification                          & NTU RGB+D~\cite{shahroudy2016ntu} \\
Comprehensive                           & VideoChat2-IT*~\cite{li2024mvbench}, FineVideo~\cite{Farré2024FineVideo} \\
    \end{tabular}
    \centering
        \caption{Video-text datasets used in the pre-training stage.}
\label{tab:ds_pretrain_video}
\end{subtable}
\label{tab:ds_pretrain}
\vspace{-1em}
\end{table}
\begin{table}[t]\scriptsize
\renewcommand{\arraystretch}{1.2}
\renewcommand{\thetable}{7}
\caption{\textbf{Summary of datasets used in the instruction tuning stage.} *IT refers to the instruction tuning data in VideoChat2.
}

\begin{subtable}{0.47\textwidth}
    \centering
    \setlength\tabcolsep{6.4pt}
    \begin{tabular}{l|l}
         task & dataset \\
\hline
General QA                        & VQAv2~\cite{goyal2017making}, GQA~\cite{hudson2019gqa}, OKVQA~\cite{marino2019ok}, VSR~\cite{liu2023visual} \\
\rowcolor{gray!15}
                                  & AI2D~\cite{kembhavi2016diagram}, ScienceQA~\cite{lu2022learn}, Chemistry Data~\cite{li2024chemvlm} \\
\rowcolor{gray!15}
\multirow{-2}{*}{Science}         & TQA~\cite{kembhavi2017you} \\

                                  & PMC-VQA~\cite{zhang2023pmc}, VQA-RAD~\cite{lau2018dataset}, VQA-Med~\cite{ben2019vqa} \\
                                  & Medical-Diff-VQA~\cite{hu2023medical}, PathVQA~\cite{he2020pathvqa}, 
 \\
\multirow{-3}{*}{Medical}         & SLAKE~\cite{liu2021slake}, PMC-CaseReport~\cite{pmc2023case} \\

\rowcolor{gray!15}
                                  & ChartQA~\cite{masry2022chartqa}, LRV-Instruction~\cite{liu2023mitigating}, PlotQA~\cite{methani2020plotqa} \\
\rowcolor{gray!15}
                                  & Unichart~\cite{masry2023unichart}, MMC-Inst~\cite{liu2023mmc}, DVQA~\cite{kafle2018dvqa} \\
\rowcolor{gray!15}
                                  & TableMWP~\cite{lu2022dynamic}, FigureQA~\cite{kahou2017figureqa}, MapQA~\cite{chang2022mapqa}  \\
\rowcolor{gray!15}
\multirow{-4}{*}{Chart}           & SciTSR~\cite{chi2019complicated}, Fintabnet~\cite{zheng2021global} \\

                                  &  CLEVR~\cite{johnson2017clevr}, MetaMath~\cite{yu2023metamath}, GeoQA+~\cite{cao2022augmented} \\
                                  & Geometry3k~\cite{lu2021inter}, GeoS~\cite{seo2015solving}, Unigeo~\cite{chen2022unigeo} \\
\multirow{-3}{*}{Mathematics}     & Super-CLEVR~\cite{li2023super}, MathQA~\cite{amini2019mathqa}\\

\rowcolor{gray!15}
                                  & Art500k~\cite{mao2017deepart}, MovieNet~\cite{huang2020movienet}, KonIQ-10k~\cite{hosu2020koniq} \\
\rowcolor{gray!15}
\multirow{-2}{*}{Knowledge}       & KVQA~\cite{shah2019kvqa}, ViQuAE~\cite{lerner2022viquae} \\

                                  & InfoVQA~\cite{mathew2022infographicvqa}, TextVQA~\cite{singh2019towards}, ArT~\cite{chng2019icdar2019} \\
                                  & CASIA~\cite{liu2011casia}, Chart-to-text~\cite{kantharaj2022chart}, COCO-text~\cite{veit2016coco} \\
                                  & CTW~\cite{yuan2019ctw}, EATEN~\cite{guo2019eaten}, ICDAR2019-LSVT~\cite{sun2019icdar} \\
                                  & ICPR MTWI~\cite{he2018icpr2018}, NAF~\cite{davis2019deep}, ReCTS~\cite{zhang2019icdar} \\
                                  & TextOCR~\cite{singh2021textocr}, LLaVAR~\cite{zhang2023llavar}, HME-100k~\cite{yuan2022syntax} \\
                                  & POIE~\cite{kuang2023visual}, SROIE~\cite{huang2019icdar2019}, ST-VQA~\cite{biten2019scene} \\
\multirow{-7}{*}{OCR}             & EST-VQA~\cite{wang2020general}, IAM~\cite{marti2002iam}\\

\rowcolor{gray!15}
Document                          & DocVQA~\cite{clark2017simple}, DocReason25k~\cite{hu2024mplug}\\

                                  & RefCOCO~\cite{kazemzadeh2014referitgame}, RefCOCO+~\cite{kazemzadeh2014referitgame}, RefCOCOg~\cite{kazemzadeh2014referitgame}  \\
\multirow{-2}{*}{Grounding}       & RD-BoxCoT~\cite{chen2023shikra} \\

\rowcolor{gray!15}
                                  & ALLaVA~\cite{chen2024allava}, LAION-GPT4V~\cite{laion_gpt4v_dataset} \\
\rowcolor{gray!15}
\multirow{-2}{*}{Conversation}    & MMDU~\cite{liu2024mmdu}, TextOCR-GPT4V~\cite{textocr-gpt4v} \\

Detection                         & Objects365~\cite{shao2019objects365}, V3Det~\cite{wang2023v3det} \\

\end{tabular}
\caption{Image-text datasets used in the instruction tuning stage.}
\label{tab:ds_sft_image}
\end{subtable}

\begin{subtable}{0.47\textwidth}
\centering
\begin{tabular}{l|l}
          task & dataset \\
\hline
                                        & ShareGPT4Video (en\&zh)~\cite{chen2024sharegpt4video}, Vript\_Chinese~\cite{yang2024vript}    \\
\multirow{-2}{*}{Detailed Caption}      & Vript~\cite{yang2024vript}, LSMDC~\cite{rohrbach2015dataset}    \\

\rowcolor{gray!15}
                                        & STAR~\cite{wu2024star}, EgoTaskQA~\cite{jia2022egotaskqa}, Mementos~\cite{wang2024mementos}            \\
\rowcolor{gray!15}
\multirow{-2}{*}{VQA}                   & TVQA~\cite{lei2018tvqa}, HiREST~\cite{Zala2023HiREST}, PerceptionTest~\cite{patraucean2024perception}     \\

Classification                          & NTU RGB+D~\cite{shahroudy2016ntu} \\
\rowcolor{gray!15}
Comprehensive                           & VideoChat2-IT*~\cite{li2024mvbench}, LLaVA-Video~\cite{zhang2024video} \\
\end{tabular}
\caption{Video-text datasets used in the instruction tuning stage.}
\label{tab:ds_sft_video}
\end{subtable}
\label{tab:ds_sft}
\vspace{-1.5em}
\end{table}
\setlength{\tabcolsep}{3pt}
\setlength{\doublerulesep}{2\arrayrulewidth}
\renewcommand{\arraystretch}{1.1}

\begin{table*}[t]
\centering
\small
\renewcommand{\thetable}{8}
\caption{\textbf{Ablation of the training data.} "New data" refers to our training dataset, which combines newly collected video-text data with InternVL2's data. Otherwise, the model is trained solely with InternVL2's data.}
\begin{tabular}{lc|ccc|ccc}
\Xhline{2\arrayrulewidth}
Model & New data & MVBench & VideoMME & MLVU & DocVQA & InfoVQA & MMB \\
\hline
InternVL2-2B & & 60.2 & 45.0 & 50.2 & 86.9 & 58.9 & 73.2 \\
PVC$_\text{InternVL2}$-2B & & 62.0 & 47.7 & 54.1 & 87.8 & 59.6 & 75.3 \\
PVC$_\text{InternVL2}$-2B & \checkmark & 69.4 & 54.5 & 63.4 & 87.7 & 59.9 & 75.8 \\
\Xhline{2\arrayrulewidth}
\end{tabular}
\label{tab:ablation_data}
\end{table*}
\setlength{\tabcolsep}{3pt}
\setlength{\doublerulesep}{2\arrayrulewidth}
\renewcommand{\arraystretch}{1.2}

\begin{table*}[ht]
\centering
\small
\renewcommand{\thetable}{9}
\caption{\textbf{Ablation of the training strategy}. The ablation study is conducted on 2B model with shortened pre-training on 10M samples. ``Unfreeze ViT \& LLM'' means unfreezing the parameters of the ViT and LLM during the pre-training stage with a lower learning rate ($\times0.1$). * InternVL2 model is trained under our ablation setting.}
\begin{tabular}{lc|ccc|ccc}
\Xhline{2\arrayrulewidth}
Model & \makecell{Unfreeze \\ ViT \& LLM} & MVBench & VideoMME & MLVU & DocVQA & InfoVQA & MMB \\
\hline
InternVL2* & & 60.9 & 44.7 & 50.0 & 86.1 & 58.7 & 74.2 \\
InternVL2* & $\checkmark$ & 61.9 & 44.9 & 50.5 & 86.4 & 59.1 & 75.1 \\
PVC$_\text{InternVL2}$ & & 60.6 & 45.5 & 53.2 & 84.8 & 57.3 & 74.4 \\
PVC$_\text{InternVL2}$ & $\checkmark$ & 62.4 & 46.7 & 55.1 & 86.7 & 58.5 & 74.9 \\
\Xhline{2\arrayrulewidth}
\end{tabular}
\label{tab:ablation_training}
\vspace{-1em}
\end{table*}

The data used in the pre-training stage are listed in Tab \ref{tab:ds_pretrain}.
All image-text data is adopted from InternVL2~\cite{internvl2}.
For video datasets, we initially utilize large-scale but mixed-quality datasets including InternVid-10M~\cite{wang2023internvid}, WebVid-10M~\cite{bain2021frozen}, TextVR~\cite{wu2025large} and OpenVid-1M~\cite{nan2024openvid} that primarily feature short video captions. To enhance the model's understanding of visual details, we further incorporate densely captioned video-text datasets of varied video lengths including MiraData~\cite{ju2024miradata}, ShareGPT4Video~\cite{chen2024sharegpt4video} and the Vript series~\cite{yang2024vript}. To improve capabilities in multi-turn conversation and visual reasoning, we employ VideoGPT+~\cite{maaz2024videogpt+}, STAR~\cite{wu2024star}, EgoTaskQA~\cite{jia2022egotaskqa}, CLEVRER~\cite{yi2019clevrer} and Mementos~\cite{wang2024mementos}. Additionally, NTU RGB+D~\cite{shahroudy2016ntu} is used to boost robustness to action recognition. Lastly, to enhance the model's holistic abilities, we utilize comprehensive datasets VideoChat2-IT~\cite{li2024mvbench} and FineVideo~\cite{Farré2024FineVideo}, which aggregate elements of multiple-choice answering, open-ended question-answering, and conversations.

Datasets used for instruction tuning are listed in Tab \ref{tab:ds_sft}.
The image-text data is also adopted from InternVL2~\cite{internvl2}.
For video-text data, low-quality datasets used in the pre-training stage are replaced by compositional high-quality datasets like LSMDC\cite{rohrbach2015dataset}, TVQA~\cite{lei2018tvqa}, HiREST~\cite{Zala2023HiREST} and LLaVA-Video~\cite{zhang2024video}.

\section{Appended Ablation Studies}
\subsection{Training Data}
To further assess the effect of the newly appended video-text data, we train a \name{}$_\text{InternVL2}$ model with the original training data of InternVL2~\cite{internvl2}.
The results are reported in Tab.~\ref{tab:ablation_data}.

\noindent\textbf{(1)} In a fair comparison with InternVL2, our PVC achieves comparable or superior performance on image benchmarks, confirming that our proposed progressive encoding effectively preserves spatial details in images. 
On video benchmarks, especially for long video tasks (e.g., VideoMME and MLVU), our PVC outperforms InternVL2 by a large margin.
This is because our PVC can process more video frames with a lower token-per-frame and better capture the temporal correlations in videos with progressive encoding.

\vspace{0.2em}
\noindent\textbf{(2)} For our PVC, incorporating the new video-text data significantly improves performance on video tasks while maintaining the performance on image tasks. 


\vspace{0.2em}
In conclusion, the significant improvements on video tasks is attributed to our PVC's strong ability to capture spatiotemporal information in videos, as well as the enhanced quality and diversity of the dataset contributed by the newly added video data.
Ablation studies in Tab.~3 and fair comparisons with InternVL2 in Tab.~\ref{tab:ablation_data} have demonstrated that our PVC attains great improvements in video understanding.
Meanwhile, current high-performing VLMs for videos, such as Qwen2-VL~\cite{wang2024qwen2} and LLaVA-Video~\cite{zhang2024video}, rely on high-quality video data, which is relatively scarce in the original dataset of InternVL2. 
To address this, we incorporate recently released open-source video data (\eg, LLaVA-Video~\cite{zhang2024video}), further elevating our PVC to state-of-the-art performance.

Moreover, the stable performance on image benchmarks further validates the robustness of our progressive encoding approach. 
While the newly introduced data is tailored to video understanding, our PVC's ability to generalize across modalities ensures that improvements on video tasks do not come at the expense of degrading performances on image tasks.
This underscores the versatility of our method in handling multi-modal inputs and highlights the complementary roles of methodology and data in driving performance improvements.

\subsection{Training Strategy}
During the pre-training phase, we unfreeze the parameters of the ViT and LLM, which differs from existing methods~\cite{liu2024visual,li2024llava,chen2024far,internvl2,bai2023qwen,wang2024qwen2}.
The ablation results in the Tab.~\ref{tab:ablation_training} empirically explain why we adopt this training strategy. 
For InternVL2, keeping ViT and LLM fixed or trainable during pre-training has minimal impact on the final performance. 
However, for our PVC, unfreezing the ViT and LLM during pre-training leads to significantly better performances.
We suppose this improvement is due to the following reasons: 
(1) Training the ViT jointly with the newly added progressive encoding module enables better capture of complementary information and minimizes redundancy.
(2) The LLM's inherent capability to process multi-frame videos is limited, especially for integrating different information extracted from the repeated frames of an image. 
Thus, additional training is needed for effective adaptation.

\subsection{Conditions of AdaLN}
We ablate the conditions used in the AdaLN layer, and the results are listed in Tab.~\ref{tab:ablation_adaln}.
Without any condition, AdaLN degrades to LN layer. 
Repeated frames of the images lack distinct timestep indicators, resulting in redundant encoding and poor performance. 
Introducing temporal embedding (TE) as the condition significantly boosts the performances. 
Adding the feature $x$ to the condition provides further improvements.
$x$ contains temporal information aggregated from previous layers, which helps better extract complementary information and minimize redundancy.

\setlength{\tabcolsep}{3pt}
\setlength{\doublerulesep}{2\arrayrulewidth}
\renewcommand{\arraystretch}{1.1}

\begin{table}[ht]
\centering
\small
\renewcommand{\thetable}{10}
\caption{\textbf{Ablation of conditions of AdaLN layers.} The ablation study is conducted on 2B model with shortened pre-training on 10M samples.}
\begin{tabular}{c|cc|cc}
\Xhline{2\arrayrulewidth}
\makecell{AdaLN \\ condition} & MVBench & VideoMME & InfoVQA & MMB \\
\hline
- & 59.2 & 46.4 & 54.0 & 73.7 \\
TE & 62.0 & 46.5 & 57.8 & 74.6 \\
$x + \text{TE}$ & 62.4 & 46.7 & 58.5 & 74.9 \\
\Xhline{2\arrayrulewidth}
\end{tabular}
\label{tab:ablation_adaln}
\end{table}

\subsection{Number of Temporal Attention Layers}
As described in Sec.3.2, we add temporal attention to the last $\tilde{L}$ layers of the ViT. 
As shown in Tab.~\ref{tab:ablation_layer}, $\tilde{L}=8$ performs better than $\tilde{L}=1$ or $\tilde{L}=4$. 
Setting $\tilde{L}=24$ (adding temporal attention to each layer) does not provide significant improvements over $\tilde{L}=8$ but increases computational overhead.
Therefore, we choose $\tilde{L}=8$, \ie, adding temporal attention to the last 8 layers of the ViT.

\setlength{\tabcolsep}{3pt}
\setlength{\doublerulesep}{2\arrayrulewidth}
\renewcommand{\arraystretch}{1.2}

\begin{table}[ht]
\centering
\small
\renewcommand{\thetable}{11}
\caption{\textbf{Ablation of the number of temporal attention layers in ViT.} The ablation study is conducted on 2B model with shortened pre-training on 10M samples.}
\begin{tabular}{c|cc|cc}
\Xhline{2\arrayrulewidth}
\makecell{\# temp. attn.\\layer ($\tilde{L}$)} & MVBench & VideoMME & InfoVQA & MMB \\
\hline
1 & 61.6 & 46.3 & 55.8 & 74.2 \\
4 & 62.1 & 46.6 & 57.3 & 74.7 \\
8 & 62.4 & 46.7 & 58.5 & 74.9  \\
24 & 62.5 & 46.7 & 58.8 & 74.8 \\
\Xhline{2\arrayrulewidth}
\end{tabular}
\label{tab:ablation_layer}
\end{table}

\section{Qualitative Results}

\begin{figure*}[t]
\renewcommand{\thefigure}{4}
\centering
    \includegraphics[width=0.96\linewidth]{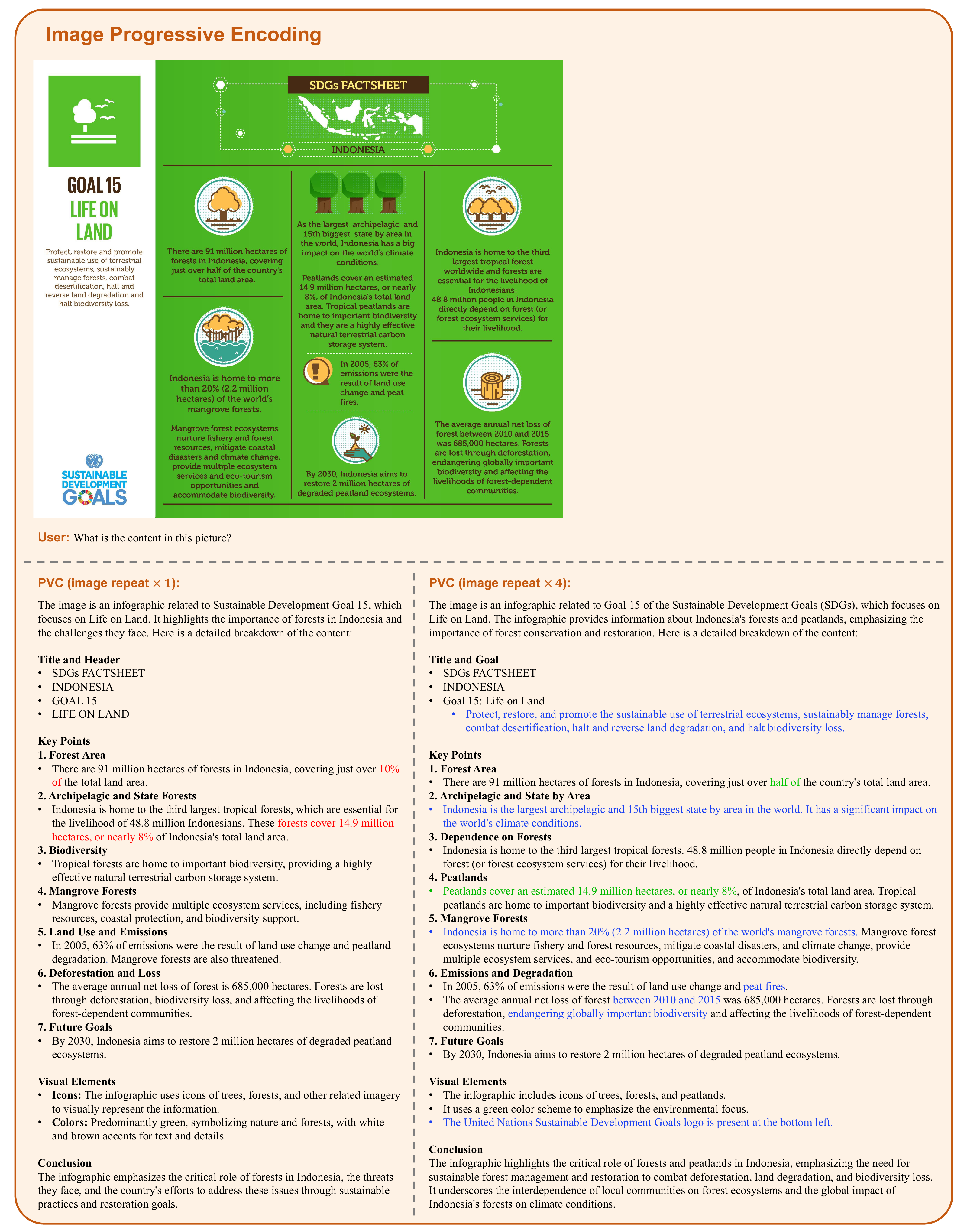}
    \caption{\textbf{PVC achieves image progressive encoding}. The image is repeated once (left) and four times (right). Supplementary contents are marked in \textcolor{blue}{blue}, incorrect contents in \textcolor{red}{red}, and corrected contents in \textcolor{green}{green}.}
    \label{fig:vis_image}
\end{figure*}

\noindent\textbf{Image Progressive Encoding.}
As shown in Fig.~\ref{fig:vis_image}, when an image is repeated as multiple frames, our PVC can extract more precise details and supplementary information from the subsequent frames.
For instance, the model extracts the title ``Goal 15 life on land'' from the first frame and corresponding detailed goals from subsequent frames. Incorrect contents, \eg ``10\% of the total land area'', are also corrected using the information extracted from subsequent frames.

\begin{figure*}[t]
\renewcommand{\thefigure}{5}
\centering
    \includegraphics[width=0.96\linewidth]{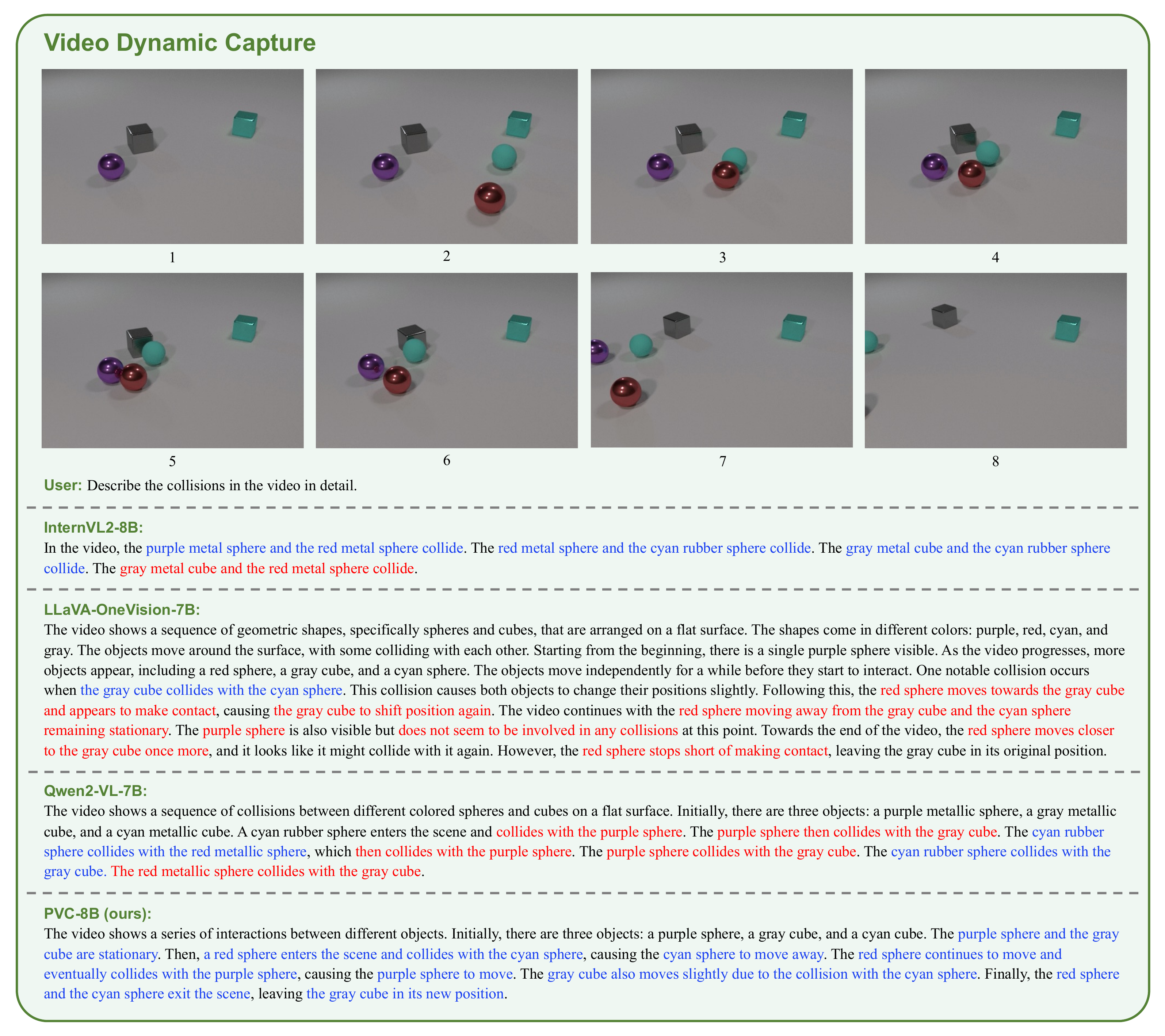}
    \caption{\textbf{PVC effectively captures spatiotemporal dynamics in videos}. Correct descriptions of the movements and interactions of the objects are marked in \textcolor{blue}{blue}, while incorrect descriptions are marked in \textcolor{red}{red}. For visualization, we select the above 8 key frames from the video, while the entire video is fed into the models.}
    \label{fig:vis_video}
\end{figure*}

\vspace{0.5em}
\noindent\textbf{Video Dynamic Capture.}
As shown in Fig.~\ref{fig:vis_video}, the video contains several consecutive collisions, testing the model's ability to capture spatiotemporal dynamics.
InternVL2~\cite{internvl2} describes each collision moment independently but lacks a description of the whole process and includes inaccuracies.
LLaVA-OneVision~\cite{li2024llava} and Qwen2-VL~\cite{wang2024qwen2} exhibit numerous errors when capturing object interactions.
In contrast, our PVC not only accurately describes the process of each collision and the associated object movements but also identifies the causal relationships between multiple collisions.

\clearpage
{
    \small
    \bibliographystyle{ieeenat_fullname}
    \bibliography{main}
}

\end{document}